\pdfoutput=1

\documentclass[11pt]{article}

\usepackage{acl}

\usepackage{times}
\usepackage{latexsym}

\usepackage{graphicx}
\usepackage[frozencache,cachedir=.]{minted}
\usepackage{amsmath}
\usepackage{amssymb}
\usepackage{textcomp}
\usepackage{booktabs}
\usepackage{float}
\usepackage{stfloats}
\usepackage{colortbl}
\usepackage{xcolor}
\usepackage{algorithm}
\usepackage{algorithmic}

\usepackage[most]{tcolorbox}

\tcbset{
  parskip/.style={
    before={\par\pagebreak[0]\parindent=0pt},
    after={\parfillskip=0pt\par},
  },
}
\newtcolorbox{resultbox}[1][]{%
    colback=black!3,
    colframe=black!3,
    notitle,
    sharp corners,
    borderline west={2pt}{0pt}{gray!80!black},
    enhanced,
    breakable,
    boxsep=0pt,
    left=4pt,right=2pt,top=2pt,bottom=2pt,
    }

\title{\texttt{MojoBench}: Language Modeling and Benchmarks for Mojo}

\author{Nishat Raihan$^{1}$, Joanna C. S. Santos$^{2}$, Marcos Zampieri$^{3}$ \\
$^{1,3}$George Mason University \quad $^{2}$University of Notre Dame \\
$^{1,3}$Fairfax, VA, USA \quad $^{2}$Notre Dame, IN, USA \\
\texttt{$^{1}$mraihan2@gmu.edu}
}

\usepackage{soul}

\DeclareMathOperator*{\argmin}{arg\,min}

\usepackage{enumitem}

\usepackage{listings}

\usepackage{minted} 
\definecolor{codebg}{rgb}{0.99,0.99,0.99}
\definecolor{hiliteColor}{rgb}{1,0.92549019607,0.6}
\definecolor{tainted}{rgb}{0,1,1}

\newmintinline{python}{fontsize=\normalsize}

\newminted[mojo]{python}{}

\newminted[PythonSourceCode]{python}{
    labelposition=topline,
    frame=single,
    numbersep=2pt,
    fontsize=\scriptsize,
    xleftmargin=5pt,
    framerule=0.5pt,
    linenos,
    escapeinside=||
}

\newcommand{\mojokw}[1]{\texttt{\textcolor[rgb]{0.125, 0.29, 0.53}{\textbf{#1}}}}  

\newminted[MojoSourceCode]{python}{
    labelposition=topline,
    frame=single,
    numbersep=2pt,
    fontsize=\scriptsize,
    xleftmargin=5pt,
    framerule=0.5pt,
    linenos,
    escapeinside=||
}

\begin{document}
\maketitle

\begin{abstract}
The recently introduced Mojo\footnote{\url{modular.com/mojo/}}  programming language (PL) by Modular,
has received significant attention in the scientific community due to its claimed 
significant speed boost over Python. Despite advancements in code Large Language Models (LLMs) across various PLs, Mojo remains unexplored in this context. To address this gap, we introduce \texttt{MojoBench}, the first framework for Mojo code generation. \texttt{MojoBench} includes HumanEval-Mojo, a benchmark dataset designed for evaluating code LLMs on Mojo, and Mojo-Coder, the first LLM pretrained and finetuned for Mojo code generation, which supports instructions in 5 natural languages (NLs). Our results show that Mojo-Coder achieves a \textit{30-35\%} performance improvement over leading models like GPT-4o and Claude-3.5-Sonnet. Furthermore, we provide insights into LLM behavior with underrepresented and unseen PLs, offering potential strategies for enhancing model adaptability. \texttt{MojoBench} contributes to our understanding of LLM capabilities and limitations in emerging programming paradigms fostering more robust code generation systems.
\end{abstract}

\section{Introduction}
LLMs demonstrate exceptional capabilities in both NLP and coding tasks, including generating executable code snippets from NL descriptions. While general-purpose models like GPT-4 \cite{gpt4omni} and the LLaMA family \cite{touvron2023llama, touvron2023llama2, dubey2024llama} exhibit strong coding abilities, task-specific models such as CodeLLaMA \cite{roziere2023code} and WizardCoder \cite{luo2023wizardcoder}, fine-tuned for code generation, often outperform them despite smaller model sizes. More recent models like CodeGemma \cite{team2024codegemma} and CodeStral \cite{mistral2024codestral} have expanded support to multiple PLs, moving beyond the predominantly Python-focused earlier models. Similar trends are also seen in existing code generation benchmarks, with limited focus on non-Python PLs.

We argue that the disproportionate focus on Python and a few other mainstream PLs overlooks the critical need to create resources for emerging and more specialized PLs. Some of these PLs, despite limited online presence, often play major roles in some sectors and domains. Mojo exemplifies this phenomenon, having rapidly ascended to the Top 100 most utilized PLs and capturing significant interest in the Machine Learning (ML) community.\footnote{\url{tiobe.com/tiobe-index/}}  
Current LLMs fail to support Mojo and similar emerging PLs, even for basic tasks like code completion. For example, as illustrated in Listing~\ref{fig:example1}, Clause 3.5 Sonnet generates \textit{Python} code instead of \textit{Mojo} code, as explicitly instructed. This glaring disparity demands immediate attention and underscores the urgent need for more inclusive, diverse PL support in LLM development.

\begin{listing}[!ht]
	\begin{PythonSourceCode*}{}
# A Mojo function to reverse a string

def reverse_string(input_string):
    return input_string[::-1]

reverse_string("hello")
	\end{PythonSourceCode*}
\caption{A Python code snippet, generated by \textit{Claude-3.5-Sonnet} when prompted to ``\texttt{write a function in Mojo to reverse a string}''.}\label{fig:example1}
\end{listing}

\noindent To address this need, we introduce \texttt{MojoBench}, a comprehensive framework for Mojo code \textit{evaluation} and \textit{generation}. Within this framework, we present the HumanEval-Mojo benchmark, designed to assess model performance on Mojo, a PL previously excluded from such evaluations. This benchmark allows us to examine state-of-the-art (SOTA) LLM performance on a PL largely unseen in training data. Mojo, having been introduced recently (2023) with a limited web presence, serves as an ideal candidate for this investigation. 

As a part of \texttt{MojoBench}, we also develop Mojo-Coder, a specialized family of Code LLMs trained for Mojo code generation from NL instructions. Our evaluations demonstrate that Mojo-Coder outperforms SOTA models such as GPT-4o \cite{gpt4omni} and CodeLLaMA \cite{roziere2023code}. Through multilingual supervised fine-tuning (SFT), Mojo-Coder supports instructions in five NLs: English, German, French, Spanish, and Bangla. 

Our main contributions with \texttt{MojoBench} are:

\begin{itemize}
    \item HumanEval-Mojo, the first benchmark designed specifically for evaluating Code LLMs on Mojo programming tasks.
    \item Mojo-Coder, a novel family of Code LLMs, pretrained and finetuned to surpass SOTA models in Mojo code generation, with support for five natural languages.
    \item The open-source release of the model, benchmark, two instruction-tuning datasets (Mojo-SFT and multilingual Mojo-mSFT), and the Mojo-Corpus to facilitate further research and development\footnote{\url{github.com/mraihan-gmu/MojoBench}}.
\end{itemize}

\noindent We use the suite of resources in \texttt{MojoBench} to address two important research questions (RQs):

\begin{itemize}
    \item {\bf RQ1:} How do LLMs perform on coding tasks in PLs that are either unseen or minimally represented in their training data, and what challenges are encountered?
    \item {\bf RQ2:} How can LLMs be effectively adapted to emerging or underrepresented PLs with limited resources?
\end{itemize}

\section{Background \& Motivation}

Mojo, introduced by Modular in 2023, is engineered for high-performance computing and machine learning. With its static typing, manual memory management, and SIMD (Single Instruction, Multiple Data) support, Mojo compiles directly to machine code, rivaling C++ performance and outpacing Python by up to 68,000 times according to \citet{deo2024performance}. At its core, the MLIR (Multi-Level Intermediate Representation) compiler framework enables advanced optimizations crucial for AI applications\footnote{\url{github.com/tairov/llama2.mojo}}.

A critical limitation of existing LLMs is the the lack of support for emerging PLs like Mojo. Most LLMs primarily serve established languages such as Python, but this oversight hinders Mojo's adoption and optimization in AI-intensive applications where its performance edge is most needed. Leading models including GPT-4 \cite{gpt4omni}, CodeLLaMA \cite{roziere2023code}, and WizardCoder \cite{luo2023wizardcoder} still lack any Mojo support, despite its impressive capabilities. This deficiency limits developers' access to AI-driven tools for enhancing productivity and optimizing Mojo implementations. A dedicated Mojo-compatible LLM as presented in \texttt{MojoBench} will help to bridge this gap and equip developers with powerful tools.

\section{Related Work}
\label{sec:rw}

While there has been no specific work on Mojo, related work has explored PL-specific models and benchmarks.

\paragraph{PL-specific Corpora} Code LLMs typically train on a combination of large NL corpora like Common Crawl \cite{CommonCrawlCorpus} and multi-PL corpora such as The Stack \cite{kocetkov2022TheStack,lozhkov2024starcoder} and CodeParrot \cite{CodeParrot}. Although PL-specific corpora are rare, they can be extracted from larger datasets, albeit through a challenging process. However, using multi-PL corpora for specific PL tasks often leads to suboptimal performance, as demonstrated by the \citeauthor{team2024codegemma}.

\paragraph{PL-specific Benchmarks} Well-established benchmarks like HumanEval \cite{chen2021evaluating}, MBPP \cite{austin2021program}, and CSEPrompts \cite{raihan2024cseprompts} primarily evaluate Python code generation. Recent extensions such as HumanEval-XL \cite{peng2024humaneval} and mHumanEval \cite{raihan2024mHumanEval} have expanded language coverage to include Java, C, C++, and others. Despite these advancements, many important PLs from the over 100 languages with substantial user bases remain underrepresented.

\paragraph{PL-specific LLMs} Most of Code LLMs, such as MagiCoder \cite{wei2023magicoder}, are tailored for Python, being fine-tuned primarily to generate Python code. PL-specific models like StarCoder-Java \cite{rathinasamy2024narrow} are exceptions. In contrast, multi-PL models are more widespread, with proprietary systems like GPT-4 \cite{gpt4omni} and the Claude family \cite{claude35} standing out as prominent examples.

\begin{figure*}[!t]
    \centering
    \includegraphics[width=0.99\textwidth]{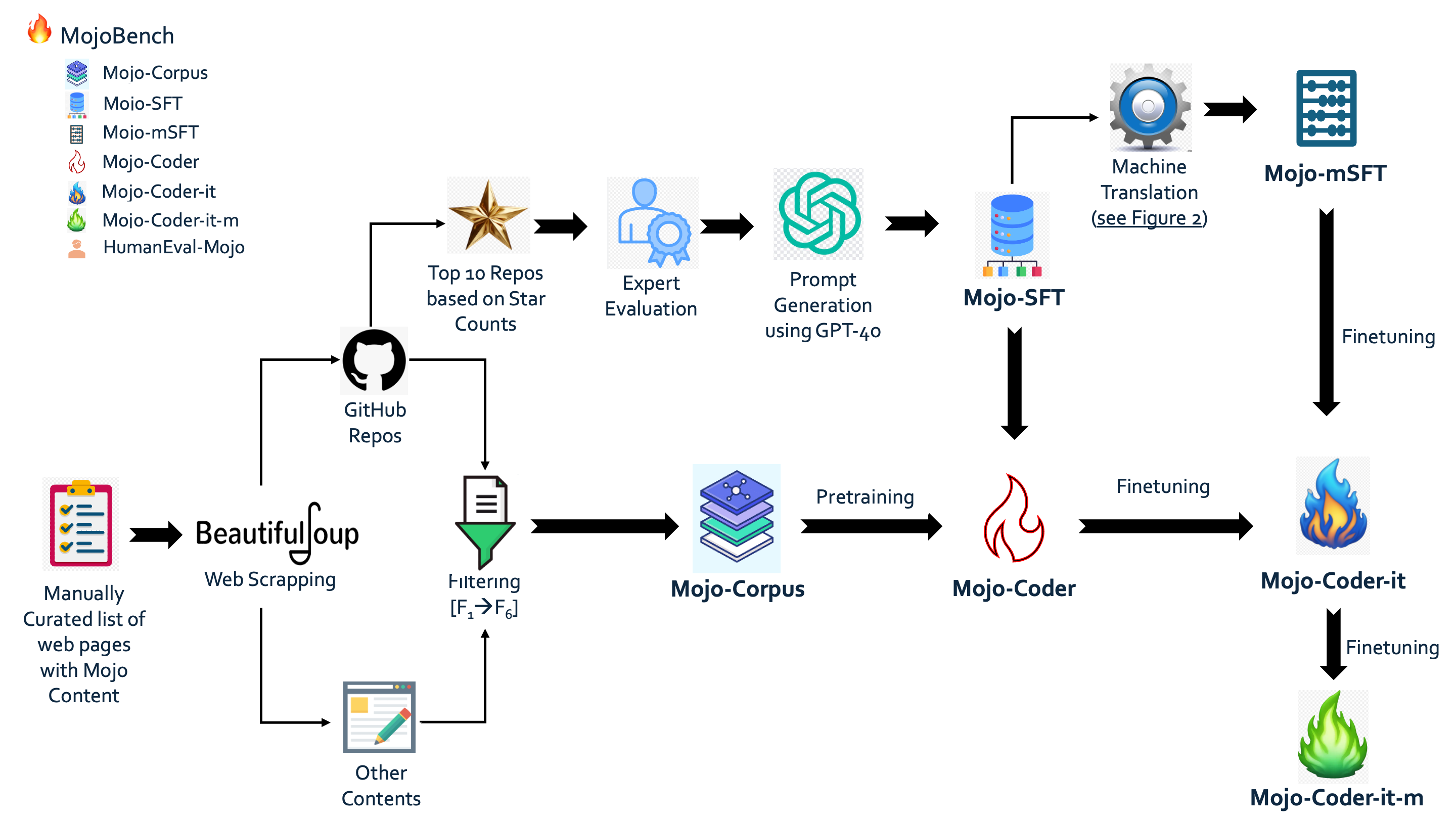}
    \caption{The complete workflow of developing \texttt{MojoBench} and all of its components. See Section \ref{sec:humaneval_mojo} for HumanEval-Mojo.}
    \label{fig:diagram1}
\end{figure*}

\section{\texttt{MojoBench}}

This section provides an overview of our \texttt{MojoBench} framework and its components. Figure \ref{fig:diagram1} shows its complete workflow.

\subsection{Mojo-Corpus: A Corpus of Mojo Code} 
\label{sec:corpus}

\texttt{MojoBench} includes a task-specific corpus. We curate data from publicly available sources, including Modular's official documentation\footnote{\url{docs.modular.com/}}, GitHub repositories, blogs, tutorials, and news articles. As discussed in the Ethical Considerations Section, we collect data from open repositories -  comprehensive list of these sources is in Appendix \ref{app:resources}. We use BeautifulSoup\footnote{\url{beautiful-soup-4.readthedocs.io/en}}, a commonly used Python library for data collection. 

Following data collection, we implement a comprehensive cleaning process to ensure corpus quality. This process applies six sequential filters ($F_1$ through $F_6$) to the initial Mojo corpus (MC), which contains 79,368,439 tokens. Table \ref{tab:transformations_tokens} displays the token count after each filtering stage. More details about these filters are provided in Appendix \ref{app:filter}.

    
    
    
    
    

\begin{table}[!ht]
\small
\setlength{\tabcolsep}{1pt} 
\centering
\begin{tabular}{cp{5.2cm}c}
\toprule
\textbf{Filter} & \textbf{Description} & \textbf{\# Tokens} \\
\midrule
\noalign{\vskip-0.8mm}
None  & All Collected Contents  & \cellcolor{teal!70} 79,368,439 \\
$F_1$   & Removes non-Apache 2.0 licensed samples. & \cellcolor{teal!50} 42,245,342 \\
$F_2$   & Removes Python-specific code snippets.       & \cellcolor{teal!40} 21,973,419 \\
$F_3$   & Ensures samples have at least 3 meaningful paragraphs. & \cellcolor{teal!30} 15,410,741 \\
$F_4$   & Removes samples with excessive internal repetition. & \cellcolor{teal!25} 10,112,466 \\
$F_5$   & Removes duplicate samples across the corpus. & \cellcolor{teal!20} 7,145,333 \\
$F_6$   & Filters non-English content using fastText.  & \cellcolor{teal!15} 6,583,948 \\
\noalign{\vskip-0.5mm}\bottomrule
\end{tabular}
\caption{Filtering all collected tokens following specific filtering criteria. For convenience, each filter is denoted by $F_i$. The right-most column shows the token count after applying each filter.}
\label{tab:transformations_tokens}
\end{table}

\subsection{{Mojo-SFT \& Mojo-mSFT}}

\texttt{MojoBench} also includes two separate instruction datasets. Mojo-SFT, comprising instructions exclusively in English, and Mojo-mSFT, encompassing instructions in Spanish, German, French, and Bangla.

\begin{figure*}[!t]
    \centering
    \includegraphics[width=0.9\textwidth]{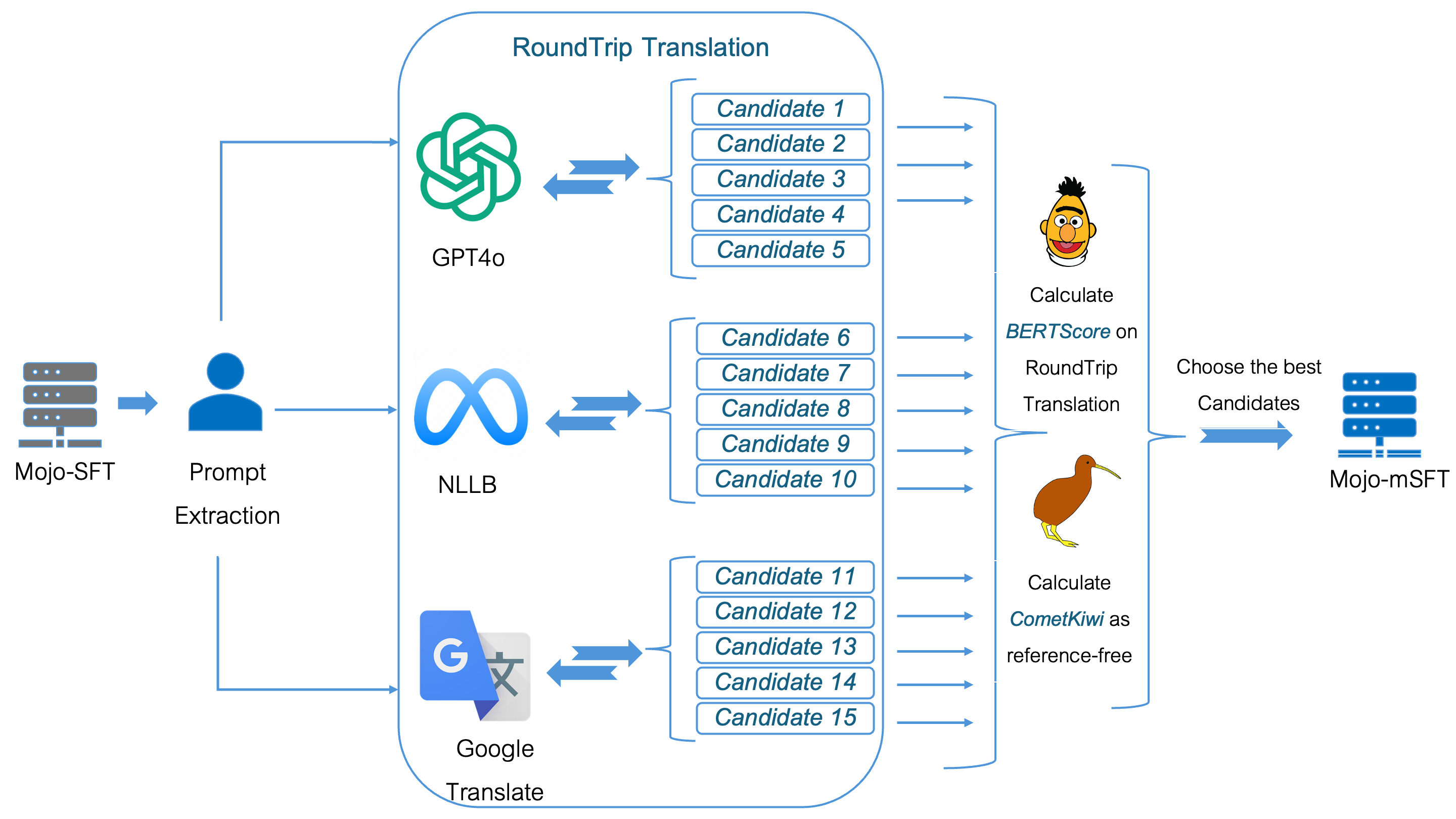}
    \caption{The workflow of compiling Mojo-mSFT from Mojo-SFT. Similar to the approach adopted by \cite{raihan2024mHumanEval}.}
    \label{fig:diagram}
\end{figure*}

\subsubsection{Mojo-SFT}

We initiate by aggregating GitHub repositories for the Mojo-Corpus, $MC^*$. Let $\mathcal{R} = \{r_1, r_2, \dots, r_n\}$ be the repositories licensed under Apache 2.0 that have undergone all transitions ($T_1$ to $T_6$) as detailed in Section \ref{sec:corpus}. Repositories are ranked by GitHub star count, $s(r_i)$, and the top 10 repositories, $R = \{r_1, r_2, \dots, r_{10}\}$, are selected such that:

\[
\forall r_i \in R, \forall r_j \in \mathcal{R} \setminus R: s(r_i) \geq s(r_j)
\]

From these, we extract \texttt{.mojo} and .\includegraphics[height=1.2em]{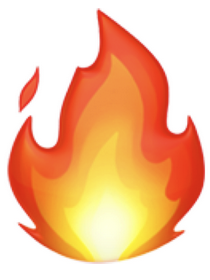} files, yielding 968 code files. To maintain consistency, we filter files by token count $\tau(f)$ as follows:

\[
\tau(f) = \begin{cases} 
1 & \text{if } 5 \leq \text{token\_count}(f) \leq 500 \\
0 & \text{otherwise}
\end{cases}
\]

where $f$ denotes a code file. Expert programmers review these, ultimately selecting 800 high-quality samples, discarding any that are buggy or too lengthy.

For prompt generation, we apply an \textit{AI-in-the-loop} approach using GPT-4o \cite{gpt4omni} via the OpenAI API, producing three paraphrases per prompt. This results in 3,200 prompt-code pairs, with iterative expert refinement yielding four distinct prompts per snippet. Table \ref{tab:mojo-sft-stats} provides summary statistics for the Mojo-SFT dataset.

\begin{table}[!t]
\small
\centering
\begin{tabular}{l r}
\toprule
\textbf{Feature} & \textbf{Statistics} \\
\midrule
Code blocks & 3200 \\
Avg (Tokens) & 216.53 \\
Median (Tokens) & 210.00 \\
Std. Dev. (Tokens) & 59.77\\
Range (Tokens) & 89 to 510 \\
Comments (Full-line) & 16,598\\
Comments (Inline) & 1,480\\
Definitions (Function) & 6,368\\
Definition (Struct) & 12 \\
\bottomrule
\end{tabular}
\caption{Mojo-SFT - Dataset Card.}
\label{tab:mojo-sft-stats}
\end{table}


\subsubsection{Mojo-mSFT}

For the multilingual dataset Mojo-mSFT, we utilize a synthetic approach across five languages, following \cite{raihan2024mHumanEval}. We employ three machine translation strategies: OpenAI's GPT-4o \cite{gpt4omni} via API, MetaAI's NLLB \cite{costa2022no}, 
and Google Translate.
Figure \ref{fig:diagram} illustrates the workflow.

To assess translation quality, we apply a dual-metric approach. BERTScore \cite{zhang2019BERTScore} measures similarity using contextual embeddings, while CometKiwi \cite{rei2023scaling} evaluates based on human-judgment-trained metrics, capturing both technical accuracy and linguistic naturalness. These metrics, discussed further in Appendix \ref{app:metric}, provide a comprehensive evaluation of translation quality. We choose the best candidate translation from a set of 15 for each prompt.

\subsection{HumanEval-Mojo}
\label{sec:humaneval_mojo}

The original HumanEval benchmark \cite{chen2021evaluating} includes 164 coding tasks, paired with test cases, and initially supported only Python. While support for a total of 43 other PLs have been provided with the works of \citet{yan2023codescope}, \citet{raihan2024mHumanEval} and \citet{peng2024humaneval}; none of them include Mojo along with a quite a few other widely used PLs.

\begin{listing}[!t]
    \centering
    \includegraphics[width=0.9\columnwidth]{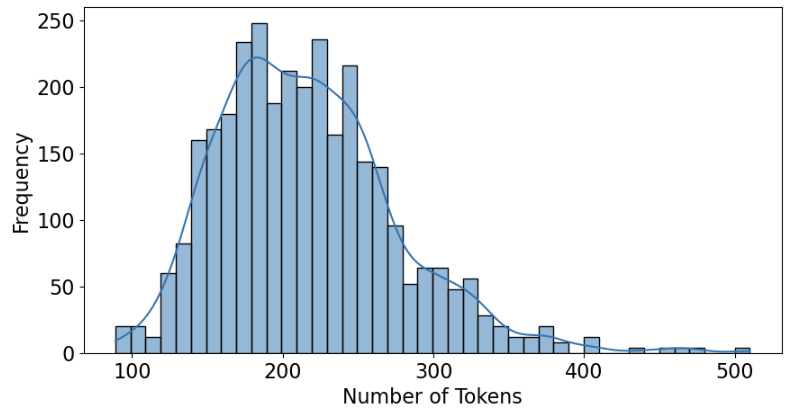}
    \caption{Code Lengths in both Mojo-SFT \& Mojo-mSFT.}
    \label{lst:hist}
\end{listing}

\begin{listing}[!ht]
    \begin{PythonSourceCode*}{}
def is_prime(n: int):
    """
    Return True if the input number n is 
    prime, else return False.
    A prime number is a number greater 
    than 1 and has no divisors other than 
    1 and itself.
    """
    \end{PythonSourceCode*}
    \caption{A sample prompt from HumanEval.}
    \label{lst:sample_1}
\end{listing}

\noindent We extend the original HumanEval for each of the 164 prompts to construct HumanEval-Mojo, an evaluation benchmark under \texttt{MojoBench}. A sample prompt is shown in Listing \ref{lst:sample_1}. We extract the docstrings manually and two human experts convert each of the Python function declarations to Mojo syntax, paired with the extracted docstrings (see Listing \ref{lst:sample_hm}). To ensure correctness, they are tested on both the local environment and the official Mojo PlayGround\footnote{\url{https://docs.modular.com/mojo/playground}}.

\begin{listing}[!ht]
    \begin{MojoSourceCode*}{}
|\mojokw{fn}| sum_squares(n: Int) -> Int:
    """
    Given an integer n, return the sum 
    of the squares of all integers from 
    1 to n (inclusive).
    """
    \end{MojoSourceCode*}
    \caption{A sample prompt from HumanEval-Mojo.}
    \label{lst:sample_hm}
\end{listing}

\noindent Finally, we provide solutions for each prompt in Mojo; hand-written by two experienced Mojo programmers. While we can not make sure that each solution is the most optimal one for the task, we do make sure that they pass all the test cases. One such solution is shown in Listing \ref{lst:sol_a}.

\begin{listing}[!ht]
    \begin{MojoSourceCode*}{}
|\mojokw{fn}| count_vowels(s: String) -> Int:
    """
    Given a string s, return the number 
    of vowels (a, e, i, o, u) in the 
    string.
    """
    |\mojokw{let}| vowels = "aeiouAEIOU"
    |\mojokw{var}| count: Int = 0
    for char in s:
        if char in vowels:
            count += 1
    return count
    \end{MojoSourceCode*}
    \caption{A sample hand-crafted solution from HumanEval-Mojo.}
    \label{lst:sol_a}
\end{listing}

\noindent Unlike \citet{yan2023codescope}, \citet{peng2024humaneval} or other benchmarks that cover more PLs, but follow an automated pipeline or use machine-generated contents, which are often prone to oversights, ours follow the original workflow, incorporating manual approach with a human-expert in the loop, ensuring better quality and more authenticity.

\subsection{Mojo-Coder}

\subsubsection{Base Model Selection} \label{sec:base_model}
Since our goal is to train a Code LLM that will be generating output in a new PL and should be able to support instructions in multiple NLs, we start by considering 3 candidates; including code-finetuned models like CodeLLaMA \cite{roziere2023code} and CodeGemma \cite{team2024codegemma}, both trained on multiple PLs and NLs and the recent version of Mistral (v0.3) \cite{mistral7b}, proficient on multiple NLs with moderate performance on non-Python PLs. Due to the limited training content, we consider the 7B variants for each.

\subsubsection{Pretraining}

Following transformations $F_1$ through $F_6$, we utilize the refined Mojo corpus $MC^*$ for additional pre-training of our selected base models.

Let $\theta_0$ represent the initial parameters of a base model. We aim to find $\theta^*$ as follows:

\begin{equation}
\theta^* = \argmin_\theta \mathcal{L}(\theta; MC^*)
\end{equation}

where $\mathcal{L}$ is the language modeling loss over the cleaned Mojo corpus $MC^*$.

\paragraph{Pre-training Setup}

Our custom training pipeline, based on the Hugging Face Transformers framework \cite{wolf2019huggingface}, includes:

\begin{equation}
\mathcal{D} = \argmin_{\mathcal{D}} \mathcal{L}(\mathcal{D}; MC^*)
\end{equation}

\begin{equation}
\tau = \argmin_{\tau} \mathcal{L}(\tau; \mathcal{T}(M_i))
\end{equation}

\begin{equation}
\mathcal{M}(\theta) = \argmin_{\theta} \mathcal{L}(\theta; \mathcal{F}(M_i))
\end{equation}

Here, $\mathcal{D}$ denotes the custom dataset from $MC^*$, $\tau$ represents the tokenizer $\mathcal{T}$, and $\mathcal{M}(\theta)$ is the pre-trained model $\mathcal{F}$ using bfloat16 precision. Both $\mathcal{T}$ and $\mathcal{F}$ derive from model family $M_i$, ensuring tokenizer-model compatibility. Training hyperparameters are detailed in Appendix \ref{app:setup}.

\begin{figure}[!h]
    \centering
    \includegraphics[width=0.9\columnwidth]{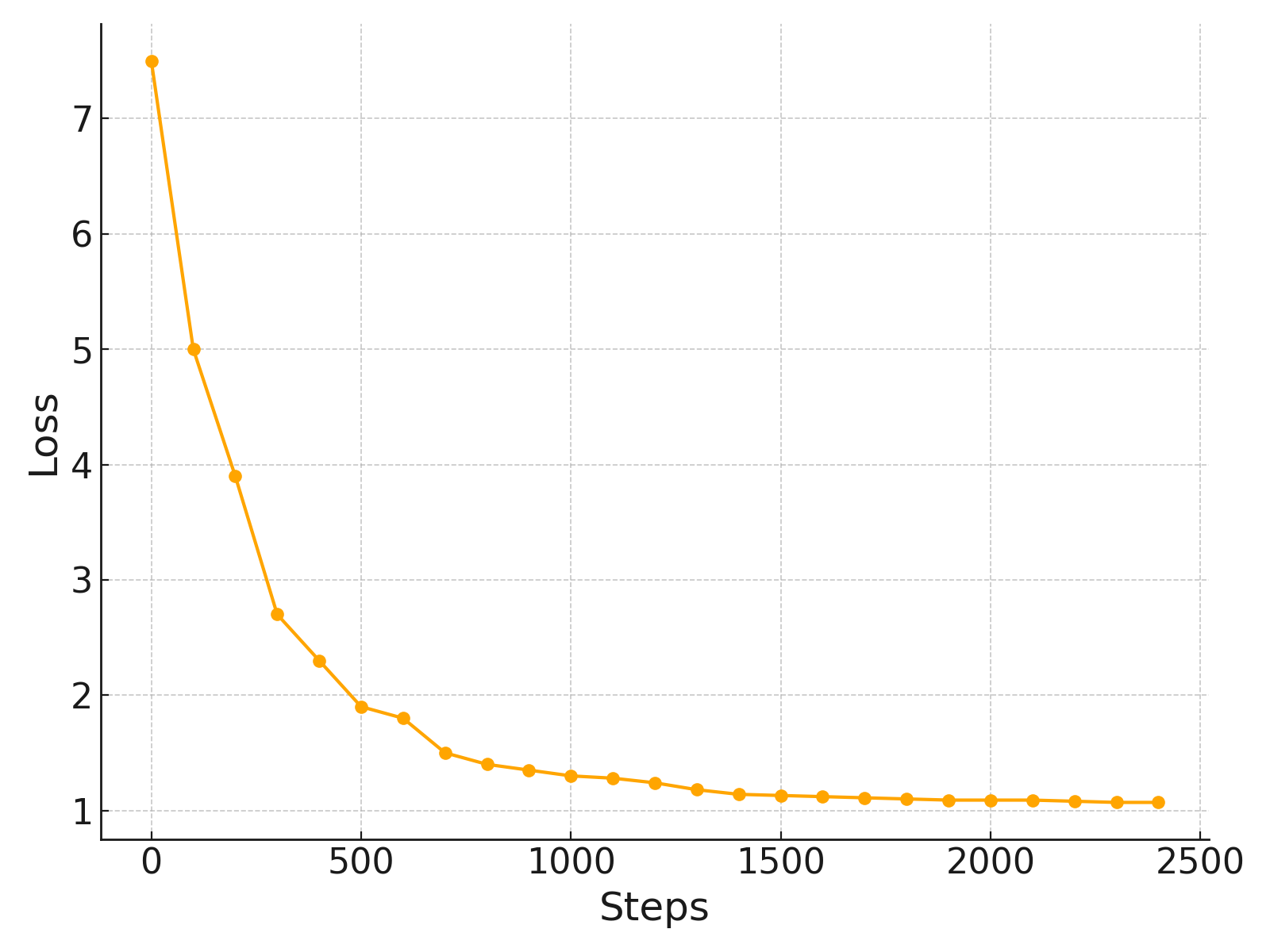}
    \caption{Step vs. Loss graph during \underline{pretraining} on Mojo-Corpus.}
    \label{fig:loss}
\end{figure}

\paragraph{Training Architecture}

The effective batch size $B_e$ is calculated as:

\begin{equation}
    B_e = B_d \times G_a \times N_d
\end{equation}

where $B_d = 32$ (per-device batch size), $G_a = 8$ (gradient accumulation steps), and $N_d$ (number of GPUs). The training steps per epoch $S_e$ are given by:

\begin{equation}
    S_e = \left\lfloor\frac{|MC^*|}{B_e}\right\rfloor
\end{equation}

where $|MC^*|$ is the total sample count in the Mojo corpus.

\paragraph{Monitoring and Evaluation}

We log the training loss $\mathcal{L}_s$ every 100 steps via a custom callback function $C(s, \mathcal{L}_s)$. Evaluation occurs every 250 steps on a 10\% subset, $MC^*_{eval} \subset MC^*$. 

Our checkpointing strategy $\Psi(s, \theta_s, \mathcal{L}_s)$ is:

\begin{equation}
\small
\Psi(s, \theta_s, \mathcal{L}_s) = 
\begin{cases}
    \text{save}(\theta_s), & \text{if } s \mod 250 = 0 \\
    \text{save}(\theta_s), & \text{if } \mathcal{L}_s = \min\left(\mathcal{L}_1, \dots, \mathcal{L}_s\right) \\
    \emptyset, & \text{otherwise}
\end{cases}
\end{equation}

Checkpoints are saved every 250 steps, retaining the model with the lowest evaluation loss. Figure \ref{fig:loss} illustrates the decreasing loss over time.

\subsubsection{Finetuning}
We finetune our three pretrained models using Mojo-SFT (English-only) and Mojo-mSFT (multilingual) instruction datasets. Implementing Low-Rank Adaptation (LoRA) \cite{hu2021lora}, a Parameter Efficient Finetuning (PEFT) technique \cite{xu2023parameter}, which optimizes memory usage while preserving performance. We choose LoRA for its ability to significantly reduce the number of trainable parameters, enabling efficient finetuning of large language models without compromising their performance \cite{hu2021lora}. For each model $M_i$ ($i \in {1,2,3}$), we define:
\begin{equation}
\small
\begin{aligned}
F_{E}(M_i) &= \text{LoRA}(M_i, Mojo-SFT, r=64) \\
F_{M}(M_i) &= \text{LoRA}(M_i, Mojo-mSFT, r=64)
\end{aligned}
\end{equation}
where $F_E$ and $F_M$ are English-only and multilingual finetuning functions, respectively. The LoRA rank $r=64$ balances parameter efficiency and model capacity.

\begin{figure}[!h]
    \centering
    \includegraphics[width=0.9\columnwidth]{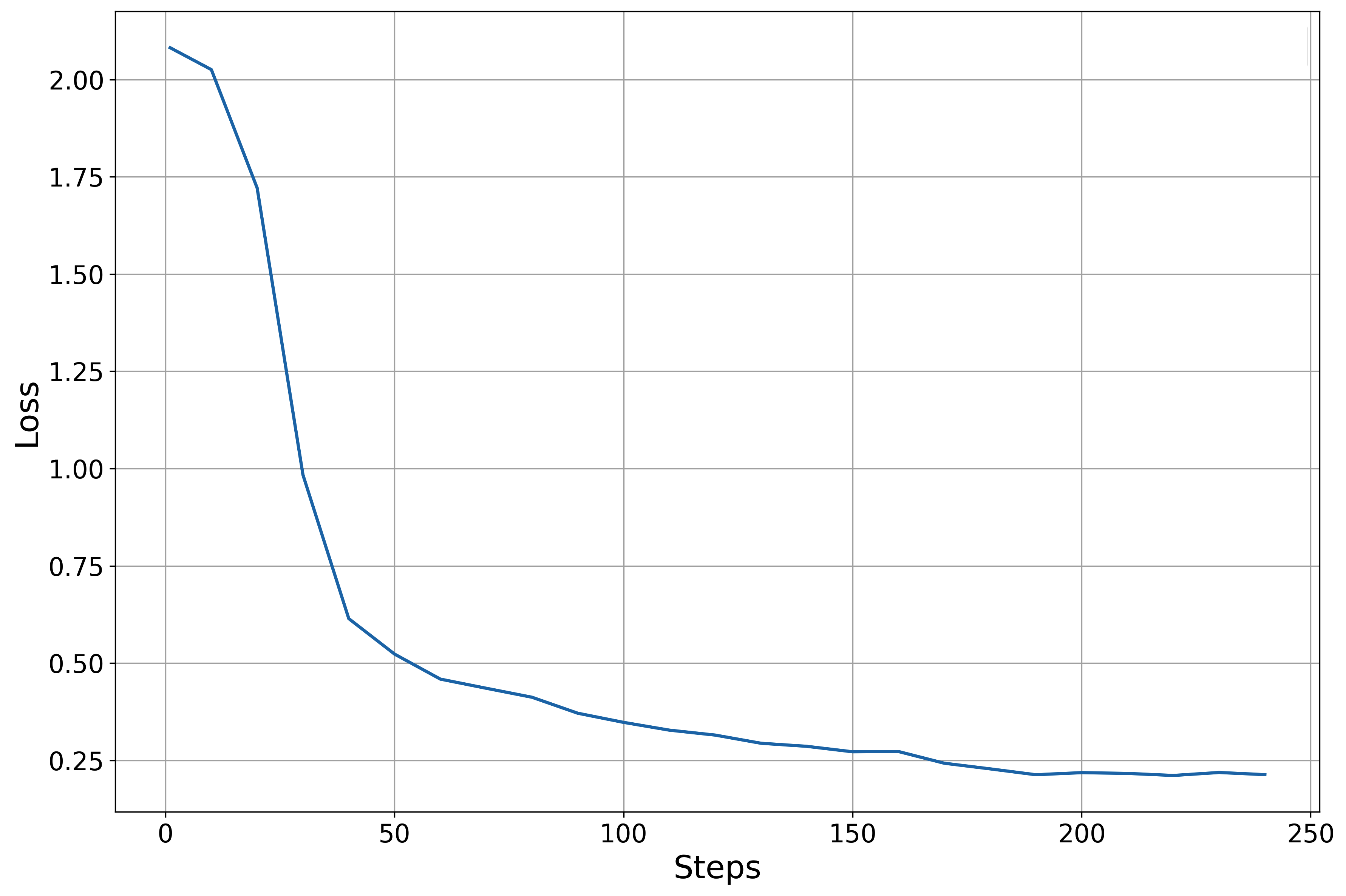}
    \caption{Step vs. Loss graph during \underline{finetuning} on Mojo-SFT.}
    \label{fig:loss2}
\end{figure}

\noindent Using Flash Attention \cite{kundu2024enhancing}, we set key parameters: 2048 token maximum sequence length, batch size of 8, 4 gradient accumulation steps, and 3 epochs. Learning rate ($5 \times 10^{-5}$), weight decay (0.02), and 10\% warm-up steps ensure stable convergence. Total training steps $T$ are calculated as:
\begin{equation}
T = \left\lfloor\frac{N}{B \times G}\right\rfloor \times E
\end{equation}
where $N$ is training samples, $B$ is batch size, $G$ is gradient accumulation steps, and $E$ is epochs. Evaluation and model saving occur every 50 steps, with the best model selected by lowest evaluation loss. A cosine learning rate scheduler and 8-bit Adam optimizer \cite{kingma2014adam} further enhance efficiency without compromising performance.
Table \ref{tab:hyperparameters2} in Appendix \ref{app:params} lists complete hyperparameters. Figure \ref{fig:loss2} demonstrates consistent loss decrease during finetuning.

\section{Results and Analysis}

\subsection{Evaluation}

\paragraph{Selecting Optimal Models} After pretraining and finetuning, we evaluate nine models: three pretrained on Mojo-Corpus, three finetuned on Mojo-SFT (English-only), and three on Mojo-mSFT (multilingual). Using HumanEval-Mojo, we select the best-performing model from each category. Table \ref{tab:candidate} compares their performance.

\begin{resultbox}
\textit{We observe poor performance in all three base models. Performance improves significantly after pretraining on Mojo-Corpus and further with Mojo-SFT finetuning. However, Mojo-mSFT finetuning leads to a slight performance decline.}
\end{resultbox}

\begin{table}[!h]
\centering
\small
\scalebox{0.99}{
    \begin{tabular}{l c}
        \toprule
        \textbf{Candidates} & \textbf{\texttt{HumanEval-Mojo pass@1}} \\
        \midrule
        CodeGemma & \cellcolor{teal!15}5.1\% \\
        CodeLLaMA & \cellcolor{teal!14}4.7\% \\
        Mistral & \cellcolor{teal!11}1.3\% \\
        \midrule
        \textbf{CodeGemma-pt} & \cellcolor{teal!36}\textbf{36.7\%} \\
        CodeLLaMA-pt & \cellcolor{teal!34}34.4\% \\
        Mistral-pt & \cellcolor{teal!23}23.1\% \\
        \midrule
        \textbf{CodeGemma-ft} & \cellcolor{teal!66}\textbf{66.4\%} \\
        CodeLLaMA-ft & \cellcolor{teal!54}54.2\% \\
        Mistral-ft & \cellcolor{teal!26}26.9\% \\
        \midrule
        \textbf{CodeGemma-mft} & \cellcolor{teal!60}\textbf{61.5\%} \\
        CodeLLaMA-mft & \cellcolor{teal!50}49.6\% \\
        Mistral-mft & \cellcolor{teal!17}17.3\% \\
        \bottomrule
    \end{tabular}
}
\caption{Candidate Model Selection after pretraining and finetuning. \textit{pt}, \textit{ft}, and \textit{mft} stand for 'pretrained' and 'finetuned,' respectively. (\textit{Pass@1}) is chosen as the accuracy metric. The darker the shade, the better the performance.}
\label{tab:candidate}
\end{table}

\paragraph{Mojo-Coder family} We release the three best performer models as the members of the Mojo-Coder family. One pretrained model, Mojo-Coder (\textit{CodeGemma-pt}), allowing practitioners to further develop models with their own instruction datasets and two finetuned models- one as English-only, Mojo-Coder-it (\textit{CodeGemma-ft}) and one multilingual Mojo-Coder-it-m (\textit{CodeGemma-mft}); both SOTA models for Mojo code generation. Appendix \ref{app:mojo_coder} illustrates some code snippets generated by Mojo-Coder.

\paragraph{Comparison} We evaluate the Mojo-Coder family of models on HumanEval-Mojo using the \textit{pass@1} metric. The results demonstrate (Table \ref{tab:comparison}) that our models significantly outperform existing state-of-the-art (SOTA) models in this benchmark.

\begin{table}[!h]
\centering
\small
\scalebox{0.99}{
    \begin{tabular}{l c c c}
        \toprule
        \textbf{Models} & Type & Param & \textbf{\texttt{Pass@1}} \\
        \midrule
        Mistral & Open & 7B & \cellcolor{teal!11}1.3\% \\
        CodeLLaMA & Open & 7B & \cellcolor{teal!14}4.7\% \\
        CodeGemma & Open & 7B & \cellcolor{teal!15}5.1\% \\
        MagiCoder & Open & 7B & \cellcolor{teal!15}7.3\% \\
        WizardCoder & Open & 34B & \cellcolor{teal!15}9.2\% \\
        Codestral & Open & 23B & \cellcolor{teal!15}9.2\% \\
        Code-Qwen & Open & 7B & \cellcolor{teal!15}9.9\% \\
        DeepSeek-Coder & Open & 33B & \cellcolor{teal!15}10.2\% \\
        GPT-4o & Close & -- & \cellcolor{teal!30}25.5\% \\
        \textbf{Mojo-Coder} & Open & 7B & \cellcolor{teal!36}\textbf{36.7\%} \\
        Claude-3.5-Sonnet & Close & -- & \cellcolor{teal!40}39.8\% \\
        \textbf{Mojo-Coder-it-m} & Open & 7B & \cellcolor{teal!60}\textbf{61.5\%} \\
        \textbf{Mojo-Coder-it} & Open & 7B & \cellcolor{teal!66}\textbf{66.4\%} \\
        \bottomrule
    \end{tabular}
}
\caption{Performance evaluation of Mojo-Coder family and other Code LLMs on HumanEval-Mojo using \textit{Pass@1}.}
\label{tab:comparison}
\end{table}

\subsection{Insights and Observations}
In this section, we describe some key insights obtained from the experiments and revisit our research questions.

\paragraph{Code LLMs \textit{vs} New PLs} 
 Existing code LLMs struggle to generate Mojo code. They often generate non-executable and buggy code snippets. Some examples are given in Appendix \ref{app:original_example}. We also include responses from other SOTA models like GPT-4o \cite{gpt4omni}, Claude-3.5-Sonnet \cite{claude35}, and WizardCoder \cite{luo2023wizardcoder}. From these results, we gather a key observation to answer our first research question:
 \begin{itemize}[leftmargin=25pt,topsep=0pt,itemsep=0pt]
     \item[$RQ_1$] \textit{How do LLMs perform on coding tasks in PLs that are either unseen or minimally represented in their training data, and what challenges are encountered?}
 \end{itemize}

\begin{resultbox}
    \textit{When prompted to write code in Mojo, the models often get the algorithm right for the tasks but the syntax wrong. This is likely due to their exposure to lots of PLs but just not Mojo. }
\end{resultbox}

Moreover, when the prompt is in any non-English language (i.e., French), these models underperform even more with mostly unexplainable and poor-quality code snippets (Appendix \ref{app:original_example}).

\paragraph{Language Modeling for Emerging Programming Languages}
This work extends beyond the finetuning approaches employed by \cite{wei2023magicoder} and WizardCoder \cite{luo2023wizardcoder} by incorporating an additional pretraining phase utilizing a corpus of 6 million tokens. While this corpus is substantially smaller than established programming language datasets \cite{lozhkov2024starcoder,CodeParrot}, our results demonstrate significant performance improvements. This outcome underscores the critical importance of acquiring domain-specific knowledge, even from limited data sources. Furthermore, our finetuning dataset, comprising only 3,200 instructions— in contrast to WizardCoder's 77,000 — accentuates the necessity of targeted, domain-specific learning. These findings directly address our second research question: 

 \begin{itemize}[leftmargin=25pt,topsep=0pt,itemsep=0pt]
     \item[$RQ_2$]\textit{How can LLMs be effectively adapted to emerging or underrepresented PLs with limited resources?}
 \end{itemize}

\begin{resultbox}
    \textit{LLMs can be effectively adapted for new or underrepresented PLs through domain-specific pretraining corpora (even a small one) and targeted instruction finetuning, prioritizing data quality over quantity to quickly capture language-specific features.}
\end{resultbox}

\subsection{Ablation Study}
For analysis purposes, we also experiment with different sizes of corpora and instruction datasets with all three models from the Mojo-Coder family, and the results further support our approaches and gathered insights.

\begin{table}[h!]
\small
\centering
\resizebox{\columnwidth}{!}{
\begin{tabular}{cccccccc}
\toprule
 & 0    & 1M & 2M & 3M & 4M & 5M & 6M \\
\midrule
0    & \cellcolor{teal!5.1} $5.1^*$  & \cellcolor{teal!12.3} 12.3 & \cellcolor{teal!14.8} 14.8 & \cellcolor{teal!16.1} 16.1 & \cellcolor{teal!21.9} 21.9 & \cellcolor{teal!30.1} 30.1 & \cellcolor{teal!36.7} $36.7^{**}$ \\
500  & \cellcolor{teal!15.3} 15.3 & \cellcolor{teal!15.6} 15.6 & \cellcolor{teal!15.3} 15.3 & \cellcolor{teal!17.9} 17.9 & \cellcolor{teal!32.1} 32.1 & \cellcolor{teal!35.5} 35.5 & \cellcolor{teal!38.8} 38.8 \\
1000 & \cellcolor{teal!19.6} 19.6 & \cellcolor{teal!21.3} 21.3 & \cellcolor{teal!20.1} 20.1 & \cellcolor{teal!25.7} 25.7 & \cellcolor{teal!37.7} 37.7 & \cellcolor{teal!39.3} 39.3 & \cellcolor{teal!41.6} 41.6 \\
1500 & \cellcolor{teal!20.3} 20.3 & \cellcolor{teal!27.8} 27.8 & \cellcolor{teal!25.3} 25.3 & \cellcolor{teal!35.6} 35.6 & \cellcolor{teal!41.2} 41.2 & \cellcolor{teal!46.7} 46.7 & \cellcolor{teal!44.5} 44.5 \\
2000 & \cellcolor{teal!22.7} 22.7 & \cellcolor{teal!34.9} 34.9 & \cellcolor{teal!36.7} 36.7 & \cellcolor{teal!38.5} 38.5 & \cellcolor{teal!42.3} 42.3 & \cellcolor{teal!51.4} 51.4 & \cellcolor{teal!53.7} 53.7 \\
2500 & \cellcolor{teal!33.3} 33.3 & \cellcolor{teal!37.2} 37.2 & \cellcolor{teal!39.8} 39.8 & \cellcolor{teal!43.1} 43.1 & \cellcolor{teal!53.4} 53.4 & \cellcolor{teal!59.4} 59.4 & \cellcolor{teal!60.1} 60.1 \\
3000 & \cellcolor{teal!42.1} 42.1 & \cellcolor{teal!43.9} 43.9 & \cellcolor{teal!49.1} 49.1 & \cellcolor{teal!56.7} 56.7 & \cellcolor{teal!55.1} 55.1 & \cellcolor{teal!60.2} 60.2 & \cellcolor{teal!64.9} 64.9 \\
3200 & \cellcolor{teal!42.3} 42.3 & \cellcolor{teal!45.3} 45.3 & \cellcolor{teal!51.2} 51.2 & \cellcolor{teal!53.4} 53.4 & \cellcolor{teal!57.9} 57.9 & \cellcolor{teal!65.1} 65.1 & \cellcolor{teal!66.4} $66.4^{***}$ \\
\bottomrule
\end{tabular}
}
\caption{Pretrained and/or finetuned on varied number of tokens (\textit{top-most row)} and/or instructions \textit{(left-most column)}. The values represent the model's \textit{Pass@1} scores. Here, $^*$ denotes CodeGemma, $^{**}$ denotes Mojo-Coder, and $^{***}$ denotes Mojo-Coder-it.}
\label{tab:ablation}
\end{table}

\noindent We derive several hypotheses from the results presented in Table \ref{tab:ablation}:

\begin{resultbox}
    \begin{itemize}[leftmargin=*, itemsep=0pt,topsep=0pt]
        \item The performance of the model exhibits a positive correlation with both the duration of pretraining and the extent of finetuning.
        \item A model subjected \textit{only} to finetuning can potentially outperform a model that has \textit{only} undergone pretraining, suggesting the critical importance of task-specific adaptation.
        \item The synergistic application of both pretraining and finetuning appears to be crucial for achieving optimal performance, indicating a complementary relationship between general knowledge acquisition and task-specific optimization.
    \end{itemize}
\end{resultbox}

\noindent It is important to note, however, that these three conclusions are drawn from a specific experimental context. Further empirical investigation across diverse datasets and model architectures would be necessary to establish the generalizability of these findings. We present these observations as promising directions for future research rather than definitive claims.

\section{Conclusion}

In this work, we presented \texttt{MojoBench}.  While most attention have been given to Python and other popular languages, with \texttt{MojoBench} we emphasize the importance of supporting underrepresented PLs, which are often newly developed or with limited online presence, fulfilling specific roles in niche and specialized domains. We focused our work on one such PL - Mojo, which, despite being recently introduced, has been gaining popularity. The paper further contributes by addressing two important RQs on LLMs and code generation. 


\texttt{MojoBench} bridges this important gap through the development of the Mojo-Coder family of models, the creation of the HumanEval-Mojo benchmark, and the compilation of two instruction datasets: Mojo-SFT and Mojo-mSFT. Our models demonstrated superior performance, even when compared to much larger proprietary models. Furthermore, we explored potential methods for adapting small Code LLMs to such emerging PLs. We expect that our methodologies and insights would encourage the research community to devote more attention to PLs that are often overlooked but nonetheless maintain a substantial user base.

\section*{Limitations}

This study introduces \texttt{MojoBench}, the first benchmark for Mojo. It includes multiple datasets, benchmarks, and models. While we prioritized quality over quantity for Mojo-Corpus, Mojo-SFT, and Mojo-mSFT, the limited availability of Mojo content on the web significantly constrained our dataset sizes. As discussed in Section \ref{sec:ethics}, in this work, only open source publicly available data was used which, in turn, reduced the datasets' scope and size. This limitation directly impacts the diversity and richness of the data used for training and evaluation, potentially affecting the models' generalization capabilities.

Furthermore, our models are confined to the $\sim$7B parameter range, a decision influenced by two primary factors. First, the relatively modest size of our datasets aligns better with smaller models, as larger models might overfit on limited data. Second, the computational intensity of pretraining favors more manageable model sizes, allowing us to iterate and experiment within our resource constraints. While this approach may limit the models' capacity compared to larger counterparts, we believe it strikes a balance between resource efficiency and model performance. 


\section*{Ethical Considerations}
\label{sec:ethics}

The datasets and models introduced in this paper, including Mojo-Corpus, Mojo-SFT, Mojo-mSFT, and the resulting language models, strictly adhere to the \href{https://www.aclweb.org/portal/content/acl-code-ethics}{ACL Ethics Policy}. We have prioritized ethical data collection practices, using only publicly available sources and respecting intellectual property rights. To ensure safety and reliability, we strongly recommend executing code generated using prompts from HumanEval-Mojo or our Mojo-Coder models in a contained virtual environment. This precaution helps prevent potential issues related to infinite execution loops, memory management, system crashes, and excessive resource consumption. We believe this approach allows researchers and practitioners to maintain a secure and controlled testing environment while confidently exploring and innovating with our resources. We remain committed to ongoing ethical evaluation and welcome community feedback to address any unforeseen concerns.

\bibliography{custom}

\clearpage

\appendix

\section{Data Sources}
\label{app:resources}

As mentioned in Section \ref{sec:corpus}, we carefully select a set of GitHub repositories with mojo code examples. Upon filtering, only the ones with Apache 2.0 licenses are kept. During extraction, we use BeautifulSoup\footnote{\url{https://beautiful-soup-4.readthedocs.io/en/latest/}}, a commonly used Python library for such purposes, to parse and extract relevant code snippets efficiently. This approach enables us to maintain consistency and accuracy in our data collection process. Table \ref{tab:mojo_repos} includes all the repository names.

\begin{table}[h] 
\centering
\small
\resizebox{0.27\textwidth}{!}{ 
\begin{tabular}{l}
\toprule
\textbf{GitHub Repositories} \\
\midrule
kojunseo/mojo-wav \\
MoSafi2/MojoFastTrim \\
mattfaltyn/mojomics \\
igorgue/firedis \\
vietanhdev/chess.mojo \\
mzaks/mojo-prefix-sum \\
HJLebbink/quine-mccluskey-mojo \\
Deftioon/Quojo \\
mzaks/mojo-sort \\
HJLebbink/sort-networks-mojo \\
alainrollejr/mocodes \\
lrmantovani10/Stable-Diffusion.mojo \\
isuckatcs/advent-of-code \\
Sharktheone/arch-mojo \\
msaelices/py2mojo \\
PriNova/MojoPkgWorkflow \\
mojopaa/menv \\
guidorice/mojo-pytest \\
joelflaig/mojo-syntax \\
Lynet101/Mojo\_community-lib \\
lsh/shims \\
mzaks/mojo-hash \\
rd4com/mojo-console-style-print \\
thatstoasty/gojo \\
tairov/llama2.mojo \\
automata/mojograd \\
andresnowak/Micro-Mojograd \\
MadAlex1997/Mojo-Arrays \\
endia-org/Endia \\
erfanzar/EasyDeL \\
StijnWoestenborghs/gradi-mojo \\
dorjeduck/momograd \\
basalt-org/basalt \\
thatstoasty/prism \\
thatstoasty/mog \\
sa-/mo-time \\
mojoto/morrow.mojo \\
Moosems/Mojo-Types \\
mzaks/mojo-trees \\
rd4com/mojo-magiclist \\
Benny-Nottonson/mojoDataStructures \\
mzaks/compact-dict \\
crisadamo/mojo-libc \\
ihnorton/mojo-ffi \\
thatstoasty/stump \\
msteele/mojo-sdl \\
rectalogic/mojo-qt \\
helehex/infrared \\
leandrolcampos/specials \\
gabrieldemarmiesse/mojo-stdlib-extensions \\
saviorand/lightbug\_http \\
taalhaataahir0102/Jpeg-Decoder \\
rd4com/mojo-learning \\
VMois/mojo-gym \\
\bottomrule
\end{tabular}
}
\caption{List of GitHub repositories for Mojo-related projects}
\label{tab:mojo_repos}
\end{table}

Additional resources include webpages with documentation, tutorials, and blogs. Again, we only consider the publicly available ones for scraping. Table \ref{tab:mojo_repos2} includes the other resources used during the compilation of Mojo-Corpus.

\begin{table}[!t]
\centering
\resizebox{0.4\textwidth}{!}{
\begin{tabular}{l}
\toprule
\textbf{Webpages that feature Mojo} \\
\midrule
\url{modular.com/mojo-programming} \\
\url{docs.modular.com/mojo-manual} \\
\url{github.com/modularml/mojo} \\
\url{codecademy.com/learn/mojo-programming} \\
\url{en.wikipedia.org/wiki/Mojo_(programming_language)} \\
\url{datacamp.com/community/tutorials/mojo-programming} \\
\url{blog.logrocket.com/getting-started-mojo-programming} \\
\url{modular.com/products/mojo} \\
\url{docs.modular.com/get-started-with-mojo} \\
\url{tutorialspoint.com/mojo_programming/index.htm} \\
\url{learnmojo.org} \\
\url{medium.com/tag/mojo} \\
\url{freecodecamp.org/news/what-is-mojo-programming-language} \\
\url{geeksforgeeks.org/mojo-programming-language} \\
\url{mojo.programming.docs.example.com} \\
\url{udemy.com/course/mojo-programming} \\
\url{coursera.org/specializations/mojo-programming} \\
\url{edx.org/course/introduction-to-mojo} \\
\url{lynda.com/Mojo-tutorials/Welcome-to-Mojo/2816042-2.html} \\
\url{khanacademy.org/computing/mojo-programming} \\
\url{academy.modular.com/mojo-programming} \\
\url{stackoverflow.com/questions/tagged/mojo-programming} \\
\url{dzone.com/articles/an-introduction-to-mojo-programming} \\
\url{realpython.com/mojo-pythonic-programming} \\
\url{pluralsight.com/courses/mojo-fundamentals} \\
\url{dev.to/t/mojo} \\
\url{hackerrank.com/domains/tutorials/mojo-programming} \\
\url{codewars.com/kata/search/mojo} \\
\url{vimeo.com/ondemand/mojoprogramming} \\
\url{ibm.com/cloud/learn/mojo-programming} \\
\url{oracle.com/mojo-programming} \\
\url{mojo.developer.com} \\
\url{guru99.com/learn-mojo-quick-guide.html} \\
\url{mojolanguage.school} \\
\url{news.ycombinator.com/item?id=31259347} \\
\url{linkedin.com/learning/mojo-programming-essentials} \\
\url{mojodojo.dev/guides/builtins/BuiltinList.html} \\
\url{mojohub.io/docs/tutorial} \\
\url{fossbytes.com/mojo-programming-tutorial} \\
\url{mozilla.org/mojo-learn} \\
\url{programiz.com/mojo-programming} \\
\url{packtpub.com/product/mojo-programming-cookbook} \\
\url{tutorialsteacher.com/mojo-programming} \\
\url{learnprogramming.com/mojo-language} \\
\url{freecodingcamp.org/mojo-programming} \\
\url{tutorialguru.com/intro-to-mojo-programming} \\
\url{coursehero.com/mojo-language-programming} \\
\bottomrule
\end{tabular}
}
\caption{List of Mojo Programming Resources}
\label{tab:mojo_repos2}
\end{table}

\section{Filltering Steps}
\label{app:filter}

\begin{enumerate}[leftmargin=*,topsep=0pt,itemsep=0pt]
    \item[$F_1$] Ensures licensing compliance by removing non-Apache 2.0 content, reducing the corpus to 42,245,342 tokens.
    
    \item[$F_2$] Focuses on Mojo-specific content by excluding Python-related snippets (e.g., \textquotesingle \textquotesingle \textquotesingle \texttt{python} or \texttt{def}), resulting in 21,973,419 tokens. This step is necessary as Python and Mojo are frequently compared online.
    
    \item[$F_3$] Enforces structural integrity by retaining samples with at least 3 code/text blocks, each containing at least 3 characters, reducing the corpus to 15,410,741 tokens.
    
    \item[$F_4$] Addresses repetition by filtering out samples with over 30\% duplicate paragraphs or 20\% duplicate characters, leaving 10,112,466 tokens.
    
    \item[$F_5$] Removes inter-sample duplicates to ensure uniqueness and reduce bias, further reducing the corpus to 7,145,333 tokens.
    
    \item[$F_6$] Applies language filtering using fastText \citep{bojanowski2017enriching}, retaining English text with a confidence threshold of 0.4, yielding the final corpus, MC*, with 6,583,948 tokens.
\end{enumerate}

\clearpage

\section{Prompt Translation and Evaluation}
\label{app:metric}

The pseudocode version of the workflow is presented in Figure \ref{fig:diagram}.

\begin{algorithm}
\caption{Prompt Translation and Evaluation Workflow}
\label{alg:translation_evaluation}
\footnotesize
\begin{algorithmic}[1]
\FOR{each extracted prompt from the HumanEval dataset}
    \FOR{each translation system}
        \FOR{each target language}
            \IF{the language is supported}
                \STATE Generate \underline{5} translated candidate prompts
                \STATE Perform back translation for each candidate
                \STATE Compute \textit{BERT\_Score} and \textit{Comet\_Kiwi} for all candidates
                \STATE Calculate the average score of \textit{BERT\_Score} and \textit{Comet\_Kiwi}
                \STATE Select the best prompt based on the highest average score
            \ELSE
                \STATE Perform back translation
                \STATE Compute only the \textit{BERT\_Score}
                \STATE Select the best prompt based on \textit{BERT\_Score}
            \ENDIF
        \ENDFOR
    \ENDFOR
\ENDFOR
\end{algorithmic}
\end{algorithm}

\subsection{Evaluation Metric 1: \texttt{BERTScore}}
\label{app:bs}

\texttt{BERTScore} leverages pre-trained BERT embeddings to quantify similarity between candidate and reference translations. For a candidate sentence $C$ and a reference sentence $R$, let $E_C$ and $E_R$ denote their respective token embedding sets. The similarity score $S(i, j)$ between tokens $i$ and $j$ is defined by the cosine similarity of their embeddings:

\begin{equation}
S(i, j) = \frac{e_{C_i} \cdot e_{R_j}}{\|e_{C_i}\| \|e_{R_j}\|}
\end{equation}

Precision $P$, recall $R$, and F1-score $F1$ are computed as follows:

\begin{align}
P &= \frac{1}{|E_C|} \sum_{e_{C_i} \in E_C} \max_{e_{R_j} \in E_R} S(i, j) \\
R &= \frac{1}{|E_R|} \sum_{e_{R_j} \in E_R} \max_{e_{C_i} \in E_C} S(j, i) \\
F1 &= 2 \cdot \frac{P \cdot R}{P + R}
\end{align}

Here, $P$ represents the average maximum similarity of each token in $C$ to any token in $R$, while $R$ denotes the average maximum similarity of each token in $R$ to any token in $C$. The $F1$ score is derived as the harmonic mean of precision and recall.

\subsection{Evaluation Metric 2: \texttt{CometKiwi}}
\label{app:ck}

\texttt{CometKiwi} (Knowledge Integration via Weighted Importance) assesses translations using human judgment scores in a reference-free context, integrating linguistic features and contextual embeddings. For a source sentence $\mathbf{x}$ and candidate translation $\mathbf{y}$, \texttt{CometKiwi} employs a neural network $\mathcal{N}$ to map inputs to a quality score $Q(\mathbf{x}, \mathbf{y})$. The network is trained on human-annotated scores $Q_{\text{human}}(\mathbf{x}, \mathbf{y})$.

The quality score is defined as:

\begin{equation}
Q(\mathbf{x}, \mathbf{y}) = f(\mathbf{E}_{\text{src}}(\mathbf{x}), \mathbf{E}_{\text{cand}}(\mathbf{y}), \mathbf{L}(\mathbf{x}, \mathbf{y}))
\end{equation}

where:
\begin{itemize}
  \item $\mathbf{E}_{\text{src}}(\mathbf{x})$ is the source sentence embedding,
  \item $\mathbf{E}_{\text{cand}}(\mathbf{y})$ is the candidate translation embedding,
  \item $\mathbf{L}(\mathbf{x}, \mathbf{y})$ represents linguistic features.
\end{itemize}

Formally, $f$ is expressed as:

\begin{equation}
f(\mathbf{E}_{\text{src}}(\mathbf{x}), \mathbf{E}_{\text{cand}}(\mathbf{y}), \mathbf{L}(\mathbf{x}, \mathbf{y})) = \mathcal{N}(\mathbf{E}_{\text{src}}, \mathbf{E}_{\text{cand}}, \mathbf{L})
\end{equation}

The network $\mathcal{N}$ is optimized to minimize the loss function $\mathcal{L}$:

\begin{equation}
\mathcal{L} = \frac{1}{N} \sum_{i=1}^{N} \left( Q(\mathbf{x}_i, \mathbf{y}_i) - Q_{\text{human}}(\mathbf{x}_i, \mathbf{y}_i) \right)^2
\end{equation}

where $N$ denotes the number of training samples.

\section{Annotator Details}

As mentioned in Section \ref{sec:humaneval_mojo}, HumanEval-Mojo involves four expert volunteers. Two of these volunteers handle the conversion of the original Python prompts into Mojo prompts, while the other two focus on crafting canonical Mojo solutions for each prompt. All four volunteers have a background in computer science, possess extensive coding experience, and are particularly skilled in the Mojo programming language.

\clearpage

\section{Experimentation Details}
\label{app:params}

\subsection{Experimental Setup}
\label{app:setup}

\paragraph{Pretraining} For the pretraining phase, we utilize a cluster of 8 NVIDIA A100 GPUs, each with 40 GB memory, provided by Lambda Labs\footnote{\url{https://lambdalabs.com/}}. This configuration yields a total of 320 GB GPU memory, complemented by 512 GB system memory and 2 TB disk space. The pretraining process spans approximately 32 hours on this high-performance cluster. 

\paragraph{Finetuning} The finetuning stage is conducted on a single NVIDIA A100 GPU with 40 GB memory, accessed through Google Colab\footnote{\url{https://colab.research.google.com/}}. This setup is augmented with 80 GB system memory and 256 GB disk space. The finetuning process requires approximately 9 hours to complete on this configuration. This more modest setup is sufficient for the task-specific adaptation of our pretrained models.

\subsection{Pretraining HyperParameters}

\begin{table}[!h]
\centering
\begin{tabular}{l c}
\toprule
\textbf{Hyperparameter} & \textbf{Value} \\
\midrule
Per device train batch size & 32 \\
Gradient accumulation steps & 8 \\
Number of training epochs & 3 \\
Learning rate & $5 \times 10^{-6}$ \\
FP16 & False \\
BF16 & True \\
Dataloader num workers & 4 \\
Gradient checkpointing & True \\
Logging steps & 100 \\
DDP find unused parameters & False \\
Max gradient norm & 1.0 \\
Warmup steps & 500 \\
Evaluation strategy & steps \\
Evaluation steps & 10,000 \\
Save strategy & steps \\
Save steps & 10,000 \\
Save total limit & 3 \\
Load best model at end & True \\
Metric for best model & loss \\
Greater is better & False \\
\bottomrule
\end{tabular}
\caption{Final set of hyperparameters, chosen empirically after several iterations of trial and error, for pretraining on the Mojo-Corpus.}
\label{tab:hyperparameters}
\end{table}

\subsection{Finetuning Hyperparameters}

\begin{table}[!h]
\centering
\label{tab:hyperparameters}
\begin{tabular}{@{}ll@{}}
\toprule
Parameter & Value \\
\midrule
Max Sequence Length & 2048 \\
Batch Size (Train/Eval) & 8 \\
Gradient Accumulation Steps & 4 \\
Number of Epochs & 3 \\
Learning Rate & 5e-5 \\
Weight Decay & 0.02 \\
Warmup Steps & 10\% \\
Optimizer & AdamW (8-bit) \\
LR Scheduler & Cosine \\
Precision & BF16 \\
Evaluation Strategy & Steps \\
Evaluation Steps & 50 \\
Save Strategy & Steps \\
Save Steps & Varies \\
Seed & 42 \\
\bottomrule
\end{tabular}
\caption{Final set of hyperparameters, chosen empirically after several iterations of trial and error, for finetuning on \underline{Mojo-SFT}.}
\label{tab:hyperparameters2}
\end{table}

\begin{table}[!h]
\centering
\label{tab:hyperparameters}
\begin{tabular}{@{}ll@{}}
\toprule
Parameter & Value \\
\midrule
Max Sequence Length & 2048 \\
Batch Size (Train/Eval) & 16 \\
Gradient Accumulation Steps & 4 \\
Number of Epochs & 3 \\
Learning Rate & 1e-5 \\
Weight Decay & 0.04 \\
Warmup Steps & 15\% \\
Optimizer & AdamW (8-bit) \\
LR Scheduler & Cosine \\
Precision & BF16 \\
Evaluation Strategy & Steps \\
Evaluation Steps & 100 \\
Save Strategy & Steps \\
Save Steps & Varies \\
Seed & 42 \\
\bottomrule
\end{tabular}
\caption{Final set of hyperparameters, chosen empirically after several iterations of trial and error, for finetuning on \underline{Mojo-mSFT}.}
\label{tab:hyperparameters2}
\end{table}

\clearpage

\onecolumn

\section{Mojo-Coder - Examples}
\label{app:mojo_coder}

\subsection{Sample Task 1}

\textit{\underline{Difficulty:}} \texttt{Easy}

\textit{\underline{Task:}} \texttt{Write a Mojo Code snippet that prints 'Hello World'}

\vspace{.4cm}

\textit{\underline{Verdict:}} \textcolor{teal}{PASSED}

\textit{\underline{Generated Response:}}

\vspace{.4cm}

\begin{listing}[!ht]
	\begin{MojoSourceCode*}{}
|\mojokw{fn}| main():
    # Print the string "Hello, World!" to the console
    print("Hello, World!")

# Call the main function to execute the program
main()
	\end{MojoSourceCode*}
    \label{fig:sol}
\end{listing}

\vspace{3cm}

\subsection{Sample Task 2}

\textit{\underline{Difficulty:}} \texttt{Easy}

\textit{\underline{Task:}} \texttt{Write a Mojo Code snippet that can detect Leap Year.}

\vspace{.4cm}

\textit{\underline{Verdict:}} \textcolor{teal}{PASSED}

\textit{\underline{Generated Response:}}

\vspace{.4cm}

\begin{listing}[!ht]
	\begin{MojoSourceCode*}{}
|\mojokw{fn}| is_leap_year(year: Int) -> Bool:
    # Check if the year is divisible by 4
    if year % 4 == 0:
        # If it's divisible by 4, check if it's not divisible by 100
        if year % 100 != 0:
            return True
        # If it's divisible by 100, check if it's divisible by 400
        elif year % 400 == 0:
            return True
        # If it's divisible by 100 but not by 400, it's not a leap year
        else:
            return False
    # If the year is not divisible by 4, it's not a leap year
    else:
        return False

# Example usage
fn main():
    let test_year = 2024
    let result = is_leap_year(test_year)
    
    if result:
        print(test_year, "is a leap year")
    else:
        print(test_year, "is not a leap year")
    
	\end{MojoSourceCode*}
    \label{fig:sol}
\end{listing}

\clearpage

\subsection{Sample Task 3}

\textit{\underline{Difficulty:}} \texttt{Medium}

\textit{\underline{Task:}} \texttt{Write a Mojo Code snippet to do Radix Sort.}

\vspace{.4cm}

\textit{\underline{Verdict:}} \textcolor{teal}{PASSED}

\textit{\underline{Generated Response:}}

\vspace{.4cm}

\begin{listing}[!ht]
	\begin{MojoSourceCode*}{}
|\mojokw{from}| math import max

fn radix_sort(inout arr: DynamicVector[Int]):
    """
    Sort the given array using radix sort algorithm.
    """
    if len(arr) <= 1:
        return

    # Find the maximum element to determine the number of digits
    max_element = arr[0]
    for i in range(1, len(arr)):
        if arr[i] > max_element:
            max_element = arr[i]

    # Determine the number of digits
    max_digits = 0
    while max_element > 0:
        max_digits += 1
        max_element //= 10

    # Perform counting sort for every digit
    for digit_place in range(max_digits):
        counting_sort(arr, digit_place)

fn counting_sort(inout arr: DynamicVector[Int], digit_place: Int):
    """
    Perform counting sort for a specific digit position.
    """
    n = len(arr)
    output = DynamicVector[Int](n)
    count = DynamicVector[Int](10, 0)

    # Count occurrences of each digit
    for i in range(n):
        index = (arr[i] // (10 ** digit_place)) % 10
        count[index] += 1

    # Change count[i] so that count[i] now contains actual
    # position of this digit in output[]
    for i in range(1, 10):
        count[i] += count[i - 1]

    # Build the output array
    i = n - 1
    while i >= 0:
        index = (arr[i] // (10 ** digit_place)) % 10
        output[count[index] - 1] = arr[i]
        count[index] -= 1
        i -= 1

    # Copy the output array to arr[], so that arr[] now contains sorted numbers
    for i in range(n):
        arr[i] = output[i]

# Example usage
fn main():
    var arr = DynamicVector[Int]()
    arr.push_back(170)
    arr.push_back(45)
    arr.push_back(75)
    arr.push_back(90)
    arr.push_back(802)
    arr.push_back(24)
    arr.push_back(2)
    arr.push_back(66)
    arr.push_back(110)
    arr.push_back(1)

    print("Original array:", arr)
    radix_sort(arr)
    print("Sorted array:", arr)
    
	\end{MojoSourceCode*}
    \label{fig:sol}
\end{listing}

\clearpage

\subsection{Sample Task 4}

\textit{\underline{Difficulty:}} \texttt{Hard}

\textit{\underline{Task:}} \texttt{Write a function in Mojo that takes a 2D grid where 1 represents land and 0 represents water. An island is a group of 1s connected horizontally or vertically. The function should return the maximum area of an island in the grid. If there are no islands, return 0.}

\vspace{.4cm}

\textit{\underline{Verdict:}} \textcolor{red}{FAILED} - Does not pass all the test cases.

\textit{\underline{Generated Response:}}

\vspace{.4cm}

\begin{listing}[!ht]
	\begin{MojoSourceCode*}{}
|\mojokw{from}| python import Python

fn max_area_of_island(grid: PythonObject) -> Int:
    """
    Find the maximum area of an island in a 2D grid.
    
    Args:
    grid: A 2D list of integers representing the grid, where 0 is water and 1 is land.
    
    Returns:
    The maximum area of an island.
    """
    
    # Get the dimensions of the grid
    rows = len(grid)
    if rows == 0:
        return 0
    cols = len(grid[0])
    
    # Initialize max_area to 0
    max_area = 0
    
    # Define a function to explore an island and calculate its area
    fn explore_island(i: Int, j: Int) -> Int:
        if i < 0 or i >= rows or j < 0 or j >= cols or grid[i][j] == 0:
            return 0
        # Mark the current cell as visited (set to 0)
        let current_area = 1
        grid[i][j] = 0
        # Explore adjacent cells
        current_area += explore_island(i + 1, j)
        current_area += explore_island(i - 1, j)
        current_area += explore_island(i, j + 1)
        current_area += explore_island(i, j - 1)
        return current_area
    
    # Iterate through each cell in the grid
    for i in range(rows):
        for j in range(cols):
            # If a land cell is found, explore its area
            if grid[i][j] == 1:
                let current_area = explore_island(i, j)
                max_area = max(max_area, current_area)
    
    return max_area

fn main():
    # Example usage
    let grid = Python.list([
        Python.list([1, 0, 0, 0, 0]),
        Python.list([0, 1, 0, 0, 1]),
        Python.list([0, 0, 1, 0, 1]),
        Python.list([0, 0, 0, 1, 0]),
        Python.list([1, 0, 0, 0, 1])
    ])
    
    let result = max_area_of_island(grid)
    print("Maximum area of an island:", result)

main()

	\end{MojoSourceCode*}
    \label{fig:sol}
\end{listing}

\clearpage

\section{Base Models vs Mojo}
\label{app:original_example}

As mentioned in Section \ref{sec:base_model}, off-the-shelf models often struggle to generate working code snippets in Mojo, when prompted. Even SOTA models like GPT-4o and Claude severely underperform when the prompt is non-English. We present some examples here reporting such issues.

\subsection{Sample Task 1}

\textit{\underline{Difficulty:}} \texttt{Easy}

\textit{\underline{Task:}} \texttt{Write a Mojo Code snippet that prints 'Hello World'}

\vspace{.4cm}

\subsubsection*{GPT-4o}
\textit{\underline{Verdict:}} \textcolor{red}{FAILED} - Parsing Error

\textit{\underline{Generated Response:}}

\begin{figure}[!h]
    \centering
    \small
    \begin{minipage}{\textwidth}
        \begin{verbatim}
fn main() {
    print("Hello World")
}
main()
        \end{verbatim}
    \end{minipage}
\end{figure}

\subsubsection*{Claude-3.5-Sonnet}
\textit{\underline{Verdict:}} \textcolor{teal}{PASSED}

\textit{\underline{Generated Response:}}
\begin{figure}[!h]
    \centering
    \small
    \begin{minipage}{\textwidth}
        \begin{verbatim}
fn main():
    print("Hello World")
        \end{verbatim}
    \end{minipage}
\end{figure}

\subsubsection*{WizardCoder}
\textit{\underline{Verdict:}} \textcolor{red}{FAILED} - Syntax Error

\textit{\underline{Generated Response:}}
\begin{figure}[!h]
    \centering
    \small
    \begin{minipage}{\textwidth}
        \begin{verbatim}
import logging

logging.info("Hello, World!")
        \end{verbatim}
    \end{minipage}
\end{figure}

\subsubsection*{CodeGemma}
\textit{\underline{Verdict:}} \textcolor{red}{FAILED} - Compilation Error

\textit{\underline{Generated Response:}}
\begin{figure}[!h]
    \centering
    \small
    \begin{minipage}{\textwidth}
        \begin{verbatim}
@Mojo( name = "hello-world")
public class HelloWorldMojo 
extends AbstractMojo {

  public void execute() throws 
  MojoExecutionException {
    getLog().info("Hello World");
  }

}
        \end{verbatim}
    \end{minipage}
\end{figure}

\subsubsection*{CodeLLaMA}
\textit{\underline{Verdict:}} \textcolor{red}{FAILED} - Compilation Error

\textit{\underline{Generated Response:}}
\begin{figure}[!h]
    \centering
    \small
    \begin{minipage}{\textwidth}
        \begin{verbatim}
<plugin>
    <groupId>com.example</groupId>
    <artifactId>helloworld
    </artifactId>
    <version>1.0.0</version>
    <executions>
        <execution>
            <phase>validate</phase>
            <goals>
                <goal>hello-world
                </goal>
            </goals>
        </execution>
    </executions>
</plugin>
        \end{verbatim}
    \end{minipage}
\end{figure}

\subsubsection*{Mistral}
\textit{\underline{Verdict:}} \textcolor{red}{FAILED} - Compilation Error

\textit{\underline{Generated Response:}}
\begin{figure}[!h]
    \centering
    \small
    \begin{minipage}{\textwidth}
        \begin{verbatim}
#include <stdio.h>

int main() {
    printf("Hello World");
    return 0;
}
        \end{verbatim}
    \end{minipage}
\end{figure}

\clearpage

\onecolumn
\subsection{Sample Task 2}

\textit{\underline{Difficulty:}} \texttt{Easy}

\textit{\underline{Task:}} \texttt{Write a Mojo Code snippet that can detect Leap Year.}

\vspace{.4cm}

\subsubsection*{GPT-4o}
\textit{\underline{Verdict:}} \textcolor{teal}{PASSED}

\textit{\underline{Generated Response:}}

\begin{figure*}[!h]
    \centering
    \small
    \begin{minipage}{\textwidth}
        \begin{verbatim}
fn is_leap_year(year: Int) -> Bool:
    if (year % 4 == 0 and year % 100 != 0) or (year % 400 == 0):
        return true
    else:
        return false

fn main():
    let year = 2024  # Change this value to test other years
    if is_leap_year(year):
        print(f"{year} is a leap year")
    else:
        print(f"{year} is not a leap year")

main()

        \end{verbatim}
    \end{minipage}
\end{figure*}

\subsubsection*{Claude-3.5-Sonnet}
\textit{\underline{Verdict:}} \textcolor{teal}{PASSED}

\textit{\underline{Generated Response:}}
\begin{figure*}[!h]
    \centering
    \small
    \begin{minipage}{\textwidth}
        \begin{verbatim}
fn is_leap_year(year: Int) -> Bool:
    return (year % 4 == 0 and year % 100 != 0) or (year % 400 == 0)

fn main():
    let current_year = 2024
    print("Is", current_year, "a leap year?", is_leap_year(current_year))

    # Test a few more years
    for year in range(2020, 2026):
        print(year, "is", "a leap year" if is_leap_year(year) else "not a leap year")

main()
        \end{verbatim}
    \end{minipage}
\end{figure*}

\subsubsection*{WizardCoder}
\textit{\underline{Verdict:}} \textcolor{red}{FAILED} - Compilation Error

\textit{\underline{Generated Response:}}
\begin{figure*}[!h]
    \centering
    \small
    \begin{minipage}{\textwidth}
        \begin{verbatim}
var leapYearApplet = {
    activate: function(args) {
        var year = args.year;
        
        // Check if the year is a leap year
        if ((year % 4 === 0) && (year % 100 !== 0) || (year % 400 === 0)) {
            Mojo.log("The year " + year + " is a leap year.");
        } else {
            Mojo.log("The year " + year + " is not a leap year.");
        }
    }
};
        \end{verbatim}
    \end{minipage}
\end{figure*}

\clearpage

\subsubsection*{CodeGemma}
\textit{\underline{Verdict:}} \textcolor{red}{FAILED} - Compilation Error

\textit{\underline{Generated Response:}}
\begin{figure}[!h]
    \centering
    \small
    \begin{minipage}{\textwidth}
        \begin{verbatim}
// Mojo code to detect leap year
int year = 2024;

// Check if the year is divisible by 4
if (year % 4 == 0) {
  // Check if the year is divisible by 100
  if (year % 100 == 0) {
    // Check if the year is divisible by 400
    if (year % 400 == 0) {
      // The year is a leap year
      print("The year " + year + " is a leap year");
    } else {
      // The year is not a leap year
      print("The year " + year + " is not a leap year");
    }
  } else {
    // The year is a leap year
    print("The year " + year + " is a leap year");
  }
} else {
  // The year is not a leap year
  print("The year " + year + " is not a leap year");
}
        \end{verbatim}
    \end{minipage}
\end{figure}

\subsubsection*{CodeLLaMA}
\textit{\underline{Verdict:}} \textcolor{red}{FAILED} - Compilation Error

\textit{\underline{Generated Response:}}
\begin{figure}[!h]
    \centering
    \small
    \begin{minipage}{\textwidth}
        \begin{verbatim}
$year = 2020;
if (date('L', mktime(0, 0, 0, 1, 1, $year)) == 1) {
    echo "It's a leap year!";
} else {
    echo "It's not a leap year.";
}
        \end{verbatim}
    \end{minipage}
\end{figure}

\subsubsection*{Mistral}
\textit{\underline{Verdict:}} \textcolor{red}{FAILED} - Compilation Error

\textit{\underline{Generated Response:}}
\begin{figure}[!h]
    \centering
    \small
    \begin{minipage}{\textwidth}
        \begin{verbatim}
#!/usr/bin/perl
use strict;
use warnings;

# Function to check if a number is prime
sub is_prime {
    my ($num) = @_;
    if ($num <= 1) {
        return 0;  # 0 and 1 are not prime
    }
    for (my $i = 2; $i * $i <= $num; $i++) {
        if ($num % $i == 0) {
            return 0;  # if divisible, not prime
        }
    }
    return 1;  # if no divisors, prime
}
        \end{verbatim}
    \end{minipage}
\end{figure}

\clearpage

\clearpage

\end{document}